\documentclass{article}

\usepackage{arxiv}

\usepackage[utf8]{inputenc} 
\usepackage[T1]{fontenc}    
\usepackage{hyperref}       
\usepackage{url}            
\usepackage{booktabs}       
\usepackage{amsfonts}       
\usepackage{nicefrac}       
\usepackage{microtype}      
\usepackage{lipsum}
\usepackage{graphicx}
\graphicspath{ {./images/} }

\title{Statistical Hand Shape Modeling from Clinical CT Scans Using Deep Learning and Implicit Skinning}

\author{
  Gökçe Güven \\
  Dept. of Computer Science and Engineering \\
  Özyeğin University \\
  İstanbul, Türkiye \\
  \texttt{gokce.guven@ozu.edu.tr} \\
  \And
  Hasan Fehmi Ateş \\
  Dept. of Artificial Intelligence and Data Engineering \\
  Özyeğin University \\
  İstanbul, Türkiye \\
  \texttt{hasan.ates@ozyegin.edu.tr} \\
  \And
  Deniz Karaşahin \\
  Osteoid Inc. \\
  İstanbul, Türkiye \\
  \texttt{deniz.karasahin@osteoid.com} \\
  \And
  Kaan Erdoğan \\
  Osteoid Inc. \\
  İstanbul, Türkiye \\
  \texttt{kaan.erdogan@osteoid.com} \\
}

\begin{document}
\maketitle

\begin{abstract}
Accurate segmentation and statistical shape modeling of hand anatomy have significant implications for medical diagnostics, ergonomics, and biomechanics. This study proposes an AI-assisted reconstruction pipeline for segmenting and analyzing hand anatomy from 1,271 elbow-to-hand (e2h-CT) computed tomography scans. A Pix2Pix-based conditional generative adversarial network is first employed to remove plaster cast and background artifacts from CT volumes. The cleaned scans are then processed in 3D Slicer to extract skin and bone masks, which are converted into closed-surface mesh models. Segmented bone meshes are used to construct skeletal representations, enabling implicit skinning to align all hand models into a standardized anatomical configuration. Subsequently, non-rigid registration is performed on the hand skin surfaces using the Geodesic Based Coherent Point Drift++ (GBCPD++) algorithm to establish point-wise correspondence across subjects. Principal Component Analysis (PCA) is then applied to the registered models to quantify anatomical shape variability. The Pix2Pix preprocessing stage achieved a Dice coefficient of 0.9856 and an IoU of 0.9720 on the held-out test set. Statistical modeling was performed on a subset of 90 scans in which the fingers were fully visible and anatomically separated. The resulting statistical shape distributions demonstrate strong agreement with the U.S. Army Anthropometric Survey (ANSUR II), supporting the anatomical validity of the reconstructed models. The proposed methodology demonstrates significant potential for advancing biomechanical modeling, ergonomic optimization, prosthetic 
design, and precision medical diagnostics.

\end{abstract}


\section{Introduction}
\label{sec:introduction}
Human hand anatomy plays a critical role in a range of domains including medical diagnostics, ergonomic design, and 
biomechanical analysis, due to its intricate structure and high inter-individual variability ~\cite{napier1993hands, moustafa2024anthropometry}. Conventional methods for analyzing hand morphology typically rely on manual measurements or semiautomated procedures, which are susceptible to operator bias, measurement inconsistencies, and limited scalability ~\cite{schmidtke2001ergonomics}.

Recent developments in deep learning, particularly convolutional neural network architectures such as U-Net~\cite{ronneberger2015u}, have significantly advanced medical image segmentation, enabling automated extraction of complex anatomical structures from volumetric imaging data with high accuracy. When integrated into anatomical reconstruction pipelines, such segmentation models enable the generation of high-fidelity 3D representations of anatomical structures, which can subsequently be normalized across subjects through skeletal pose alignment, enabling consistent anatomical comparisons at the population level.

Previous studies have proposed several approaches for constructing individualized hand models from imaging data. These include deformation based on surface anatomy cues~\cite{handsurface}, joint estimation from motion capture data~\cite{handcapture}, and CT-based modeling of deformable hand geometries across multiple poses~\cite{handdeformable}. While these approaches have successfully produced subject-specific mesh representations and skeletal structures, they often involve extensive manual intervention or rely on resource-intensive workflows that limit their applicability in large-scale or clinical contexts~\cite{handposture}. Furthermore, clinical CT datasets frequently contain artifacts such as plaster casts and background structures that complicate automated 
anatomical reconstruction, a challenge that has received limited attention in prior work.

A key challenge in constructing population-level hand models from clinical CT data is the variability in hand pose across subjects, which must be normalized prior to meaningful statistical comparison. Skinning techniques are essential for this purpose. Traditional explicit skinning methods rely on discrete mesh deformation driven by underlying skeleton movements, often resulting in visual artifacts and unnatural distortions~\cite{kavan2008geometric, jacobson2011bounded}. In contrast, implicit skinning approaches represent anatomical surfaces as continuous implicit functions, effectively overcoming these limitations by smoothly 
interpolating surface deformations and preserving anatomical accuracy~\cite{vaillant2013implicit}. Geodesic voxel binding further 
offers artifact-minimizing benefits over traditional skinning approaches by leveraging voxelized geodesic proximity~\cite{dionne2013geodesic}; although originally developed for production character meshes, such geometric principles are directly applicable to anatomical mesh deformation in medical contexts~\cite{orthoauto}.

In parallel, statistical shape modeling---particularly via PCA---provides a compact representation of population-level anatomical 
variability and supports quantitative comparisons across individuals~\cite{cootes1995active, ambellan2019statistical}. External 
validity is commonly assessed against independent anthropometric references; the 2012 U.S. Army Anthropometric Survey (ANSUR~II)~\cite{gordon2015ansur} provides normative hand measurement statistics for benchmarking learned models against population-level morphology.

To address these challenges, this study proposes an AI-assisted pipeline for generating posture-normalized statistical hand shape models from elbow-to-hand computed tomography (e2h-CT) scans. The proposed framework integrates deep learning-based preprocessing, anatomical mesh extraction, skeletal fitting, pose normalization, and statistical shape modeling within a unified pipeline. A Pix2Pix-based conditional generative model~\cite{pix2pix} is first employed to remove plaster and background artifacts from CT volumes. The cleaned volumes are subsequently processed in 3D Slicer~\cite{slicer} to extract skin and bone masks, which are converted into closed-surface mesh models. Bone meshes are then used to construct skeletal structures and perform pose normalization using implicit skinning techniques. Following pose alignment, non-rigid registration is applied using the GBCPD++ algorithm~\cite{gbcpd} to establish point-wise correspondence across subjects, enabling robust statistical analysis using Principal Component Analysis.

The main contributions of this study are:
\begin{itemize}
    \item Development of an AI-assisted CT processing pipeline combining Pix2Pix-based artifact removal with 3D Slicer-based skin and bone mesh extraction from clinical e2h-CT data.
    \item Application of implicit skinning and skeletal fitting to achieve anatomically consistent pose normalization across subjects.
    \item Construction of posture-normalized statistical hand shape models and quantitative analysis of anatomical variability using PCA.
    \item Quantitative validation against the ANSUR~II anthropometric dataset to assess the anatomical fidelity of the reconstructed models.
    \item Demonstration of the framework's applicability in clinical diagnostics, ergonomic assessment, prosthetic design, and 
    biomechanical modeling.
\end{itemize}

\section{Related Work}

\subsection{AI-Driven Segmentation from CT Scans}

Recent advancements in artificial intelligence (AI) have notably improved anatomical segmentation, particularly bone segmentation 
from computed tomography (CT) images, significantly impacting clinical diagnostics and treatment planning. U-Net-based 
architectures have become the standard for medical image segmentation due to their robust hierarchical feature extraction 
capabilities~\cite{ronneberger2015u}. For instance, Wasserthal et al. introduced TotalSegmentator, achieving highly accurate bone 
and soft tissue segmentation from whole-body CT scans with a Dice similarity coefficient of 0.943, demonstrating significant potential for clinical utility~\cite{wasserthal2022totalsegmentator}.

Further advancements include automated bone segmentation of specific anatomical regions, such as pelvic bones and 
femurs~\cite{hemke2020deep}. Lindgren et al. employed deep learning methods for automated femur segmentation, achieving a Dice similarity coefficient of 0.990, demonstrating exceptional accuracy and clinical applicability~\cite{lindgren2021automated}.

Despite these advances in general anatomical segmentation, dedicated methods for hand and wrist CT segmentation remain relatively 
underexplored, with most existing approaches relying on semi-automated or threshold-based techniques rather than end-to-end 
deep learning pipelines~\cite{orthoauto}. These advances motivate the integration of deep learning-based preprocessing into anatomical reconstruction pipelines for scalable processing of large clinical CT datasets with minimal manual intervention.

\subsection{Skinning Algorithms}

Realistic skinning algorithms are essential for simulating anatomical deformation driven by skeletal motion, which is critical 
for biomechanical analysis and ergonomic optimization. Traditional explicit skinning methods, such as linear blend skinning and dual 
quaternion skinning, are widely utilized but can suffer from unrealistic artifacts and unnatural deformations due to linear 
vertex transformations~\cite{kavan2008geometric, jacobson2011bounded}. Rhee et al. proposed a method that extracts anatomical creases from palm images using tensor voting, estimates underlying joint positions, and adapts a generic 3D hand model using RBF deformation to generate personalized hands with skinning-ready topologies~\cite{handsurface}. Kurihara and Miyata developed an 
example-based deformable hand model using multi-pose CT imaging, where joint centers were estimated from inter-bone transformations 
and surface meshes were aligned using pose-space deformation with RBF refinement~\cite{handdeformable}.

Recently developed implicit skinning techniques, employing continuous implicit representations, effectively address these 
limitations by providing smoother, artifact-free deformation with improved anatomical accuracy. Notably, implicit skinning introduced by Vaillant et al. incorporates real-time deformation and contact modeling, enhancing realism and 
interactivity~\cite{vaillant2013implicit}. Similarly, geodesic voxel binding offers artifact-minimizing benefits over traditional 
skinning approaches by leveraging voxelized geodesic proximity~\cite{dionne2013geodesic}. While originally developed for 
production character meshes, the geometric principles underlying this approach are directly applicable to anatomical mesh deformation in medical contexts, as demonstrated in the present work.

\subsection{Skinned Statistical Anatomical Models}

Statistical shape modeling methods, such as Principal Component Analysis (PCA), are widely utilized for capturing anatomical 
variability~\cite{cootes1995active, ambellan2019statistical}. Advanced methodologies, including 3D Morphable Models (3DMMs), 
enable generation of statistically plausible anatomical shapes, particularly effective in modeling faces and other articulated 
anatomical structures~\cite{blanz1999morphable}. Bayesian estimation techniques in computational anatomy have advanced 
anatomical template generation by statistically capturing population-level shape variations~\cite{joshi2004bayesian}.

Yang et al. constructed a posture-invariant statistical shape model (SSM) of the hand by applying articulated skeleton 
registration and linear blend skinning to correct for pose variability across 3D scans, with PCA-based dimensionality 
reduction providing compact and generalizable shape spaces~\cite{handposture}. However, their approach relied on 
controlled 3D scan acquisitions with anatomically separated fingers, limiting applicability to clinical CT data which 
frequently contains artifacts and incomplete field-of-view coverage. The present work addresses these limitations by 
incorporating deep learning-based artifact removal and operating directly on retrospective clinical CT volumes.

Whole-body statistical shape models such as SMPL~\cite{loper2015smpl} and BOSS~\cite{boss2022} demonstrate 
the scalability of PCA-based approaches to complex articulated anatomical structures, providing methodological foundations 
applicable to region-specific models such as the hand. Collectively, these advances demonstrate significant potential 
for clinical decision-making, ergonomic optimization, prosthetic development, and personalized medical treatments.

Despite these advances, existing approaches for constructing statistical hand models often face several practical limitations. 
Many methods rely on controlled scanning setups, motion capture data, or extensive manual intervention during model construction 
and alignment. In addition, clinical CT datasets frequently contain artifacts such as plaster casts, background structures, 
or incomplete field-of-view coverage, which complicate automated anatomical reconstruction. These challenges limit the scalability 
of current pipelines for large clinical datasets. Consequently, there remains a need for robust reconstruction frameworks that 
integrate modern AI-based preprocessing with anatomically consistent pose normalization and statistical modeling techniques. 
The approach proposed in this work aims to address these limitations by combining deep learning-based artifact removal 
with mesh extraction, skeletal alignment, and statistical shape analysis within a unified AI-assisted pipeline.

\section{Dataset}

\subsection{Cohort and Composition}

A total of 1,271 elbow-to-hand CT (e2h-CT) volumes were retrospectively collected from hospital imaging archives. This study was conducted in accordance with the Declaration of Helsinki. All CT volumes were fully de-identified prior to analysis; informed consent was waived due to the retrospective nature of the data collection.

The dataset was grouped according to (i) whether the fingers were included in the scan field-of-view (FOV) and (ii) whether 
a plaster cast was present at the time of imaging. The resulting four acquisition conditions were: fingers included 
in FOV with plaster cast ($n = 115$), fingers included in FOV without cast ($n = 246$), fingers excluded from FOV with 
plaster cast ($n = 453$), and fingers excluded from FOV without cast ($n = 457$), resulting in a total of 1,271 CT 
volumes. In this study, ``fingers excluded'' indicates that distal digits were not covered by the prescribed ROI/FOV 
during acquisition. The dataset reflects real clinical acquisition conditions and therefore includes variability in 
scan orientation, field-of-view coverage, and the presence of immobilization devices such as plaster casts, providing 
a realistic evaluation setting for the proposed reconstruction pipeline.

\subsection{Data Preparation for Pix2Pix}

To construct paired training data for the Pix2Pix model, the original CT volumes were first resampled to an in-plane resolution of $1024 \times 1024$ pixels with an isotropic spacing of 0.5~mm using the SimpleITK framework. The resampled volumes were then exported as axial 2D slices in JPEG format.

For each slice, a corresponding binary ground-truth mask was manually generated to isolate the hand and plaster regions from the surrounding background structures. Manual annotations were performed by trained operators using region selection and brush tools in Adobe Photoshop to accurately delineate the anatomical region of interest.

In total, 3,038 paired CT slices were used in the Pix2Pix training pipeline: 2,430 for training, 304 for validation, and 304 for testing. The validation set was used for model selection and hyperparameter tuning, whereas the test set was reserved exclusively for final performance evaluation.

\subsection{Dataset Partitioning for Statistical Shape Modeling}

Of the 1,271 CT volumes, 361 scans contained fingers fully visible within the field of view. Following mesh extraction and quality assessment, scans with fused digits or incomplete segmentation were excluded, yielding a final subset of 90 hand meshes retained for PCA-based statistical shape analysis. Of the 361 scans with fingers in FOV, 90 met strict anatomical quality criteria — fully separated, non-fused digits — required for valid surface correspondence and PCA. This proportion reflects the inherent challenges of retrospective clinical data, where patient positioning and pathological conditions limit the subset available for population-level shape analysis. Table~\ref{tab:dataset_split} summarizes the dataset composition across all processing stages.

\begin{table}[h]
\centering
\caption{Dataset composition across processing stages.}
\label{tab:dataset_split}
\begin{tabular}{lc}
\hline
\textbf{Stage} & \textbf{N} \\
\hline
Total CT volumes & 1,271 \\
\quad Fingers in FOV & 361 \\
\quad\quad With plaster cast & 115 \\
\quad\quad Without cast & 246 \\
\quad Fingers excluded from FOV & 910 \\
Retained for PCA analysis & 90 \\
\hline
\end{tabular}
\end{table}

\section{Method}

The proposed methodology presents a comprehensive AI-assisted pipeline for the segmentation and analysis of hand anatomy from elbow-to-hand computed tomography (e2h-CT) scans, comprising the following four principal components:

\begin{enumerate}
    \item \textbf{Pix2Pix-based preprocessing:} Detection and removal of plaster cast and background artifacts through a learned image-to-mask translation model.
    \item \textbf{3D Slicer-based mesh extraction:} Generation of skin and bone closed-surface meshes from the artifact-reduced CT volumes.
    \item \textbf{Skeletal fitting and implicit skinning:} Alignment of hand meshes into a standardized anatomical pose using implicit deformation techniques.
    \item \textbf{Statistical shape modeling with PCA:} Quantification of population-level anatomical variability through principal component analysis of registered hand surfaces.
\end{enumerate}

\subsection{Pix2Pix-Based Preprocessing and Plaster Removal}

The preprocessing stage employs a Pix2Pix-based conditional generative adversarial network (cGAN) to detect plaster casts and non-anatomical background artifacts in elbow-to-hand CT volumes. The Pix2Pix framework learns a mapping between input CT slices and corresponding binary masks that identify artifact regions, including plaster casts and non-anatomical background structures.

The model follows the Pix2Pix architecture proposed by Isola et al.~\cite{pix2pix}, where the generator adopts a U-Net-based 
encoder--decoder structure with skip connections to preserve spatial information. The discriminator is implemented as a PatchGAN classifier that evaluates local image patches to distinguish between generated and ground-truth masks.

The model was trained on axial CT slices with an input resolution of $1024 \times 1024$ pixels. Training was performed for 200 epochs (100 epochs at 
the initial learning rate followed by 100 epochs with linear decay) using the Adam optimizer with a learning rate of $2 \times 10^{-4}$, $\beta_1 = 0.5$, and a batch size of 1. The loss function combines adversarial loss with an L1 reconstruction term, where the adversarial component encourages realistic mask generation and the L1 term promotes pixel-wise structural fidelity with respect to the manually annotated ground-truth masks.

Once trained, the model is applied to all CT slices to predict artifact masks, which are then used to suppress plaster casts and background regions, producing artifact-reduced CT volumes for downstream processing.

\subsection{3D Slicer-Based Bone and Skin Mesh Extraction}

Following Pix2Pix-based preprocessing, skin and bone representations were extracted from the cleaned CT volumes using 3D Slicer~\cite{slicer}.

For \textbf{skin extraction}, air-connected voxels at the image boundary were first identified, and the largest complementary body component was retained to obtain an external soft-tissue mask.

For \textbf{bone extraction}, intensity thresholding was applied within the range of 150--3000~HU to isolate osseous structures, followed by connected-component filtering. Small connected components containing fewer than 80 voxels were removed to suppress noise while preserving the major hand and distal forearm bones.

\begin{figure}[ht]
    \centering
    \includegraphics[width=0.5\linewidth]{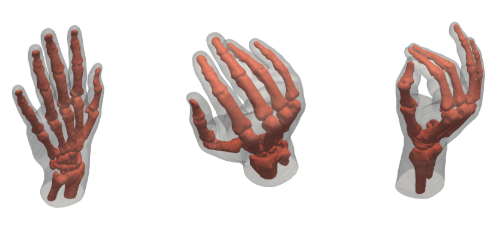}
    \caption{Example skin and bone meshes extracted from Pix2Pix-cleaned elbow-to-hand CT volumes using the proposed 3D Slicer-based segmentation workflow. The semi-transparent surface represents the skin mesh, while the internal structures correspond to the segmented bone models used for subsequent skeletal fitting and implicit skinning.}
    \label{fig:skin_bone_meshes}
\end{figure}

In both cases, the resulting binary labelmaps were converted into closed-surface mesh models using the 3D Slicer closed-surface conversion pipeline with FlyingEdges-based surface extraction and a smoothing factor of 0.25. The extracted skin and bone meshes serve as the geometric inputs for downstream skeletal fitting and implicit skinning, as illustrated in Fig.~\ref{fig:skin_bone_meshes}.

\subsection{Skeletal Fitting and Implicit Skinning}

To ensure anatomically consistent comparisons across individuals, the extracted bone meshes were used to derive simplified skeletal models, which were then aligned into a standardized pose through a skeletal fitting and implicit skinning pipeline.

\begin{figure}[ht]
\centering
\includegraphics[width=0.2\linewidth]{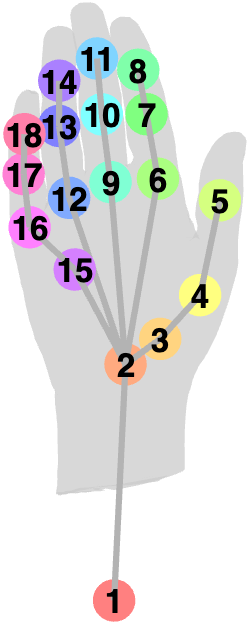}
\caption{Manual placement of skeletal landmarks used to construct the articulated hand skeleton for implicit skinning. Colored markers indicate joint centers and bone axis endpoints defining the kinematic hierarchy. The numbering corresponds to the hierarchical joint structure used in the skeletal model.}
\label{fig:skeleton_landmarks}
\end{figure}

The process begins with the construction of a simplified anatomical skeleton derived from key landmarks defined on the segmented bones. Key anatomical landmarks, including joint centers and bone axis endpoints, are manually defined by an operator to construct the articulated skeletal model, as illustrated in Fig.~\ref{fig:skeleton_landmarks}. The skeletal hierarchy follows the anatomical structure of the hand, where each finger is represented as a kinematic chain connected to a central palm structure, and each bone 
corresponds to a rigid transformation enabling mesh deformation through skeletal motion.

Skinning weights define the influence of each bone on surrounding mesh vertices. Each vertex was associated with the bone exerting dominant influence in its local region, partitioning the mesh into regions corresponding to individual bones. This enables skeletal transformations to propagate smoothly across the mesh during pose normalization, providing the initial deformation required for implicit skinning.

Once the skeleton is established, implicit skinning is employed to bind the skin surface to the underlying skeletal structure. Unlike traditional linear blend skinning, implicit skinning treats the geometry as a continuous scalar field, enabling smooth, artifact-free deformation~\cite{vaillant2013implicit}, particularly in highly articulated regions such as the wrist and finger joints.

\subsection{Statistical Shape Modeling with PCA}

All hand skin surfaces were non-rigidly registered to a common template using the Geodesic Based Coherent Point Drift++ (GBCPD++) algorithm~\cite{gbcpd} prior to Principal Component Analysis (PCA). This registration establishes point-wise correspondence across surfaces, which is a prerequisite for accurate PCA computation. GBCPD++ enables alignment of complex anatomical geometries while preserving surface topology and subject-specific features, ensuring that shape variation is analyzed under consistent spatial 
mappings.

The aligned hand models were then statistically analyzed using PCA~\cite{cootes1995active, ambellan2019statistical}, capturing the primary modes of anatomical variation across the dataset. The first principal component explained 44.0\% of the total variance, followed by 25.8\% for the second component and 12.9\% for the third component, with the first three components jointly accounting for 82.6\% of total variance. These components primarily correspond to variations in overall hand size (PC1), palm width and finger length proportions (PC2), and fine-scale shape asymmetries (PC3). Quantitative validation was performed by comparing key anthropometric measurements derived from the mean shape and PCA-derived shape range against normative values from the ANSUR~II dataset~\cite{gordon2015ansur}.

\section{Experimental Setup}

To quantitatively evaluate the Pix2Pix preprocessing stage, the predicted masks were compared against manually annotated ground-truth masks using standard segmentation metrics: Dice Similarity Coefficient (DSC), Intersection over Union (IoU), precision, recall, and pixel-wise accuracy. Image similarity was additionally assessed using mean absolute error (MAE) and structural similarity index (SSIM). All metrics were computed on the held-out test set using Python-based analysis scripts.

\section{Results}

The proposed AI-assisted pipeline was evaluated across four stages: Pix2Pix-based artifact removal, 3D Slicer-based mesh extraction, implicit skinning alignment, and PCA-based statistical shape modeling. Quantitative metrics and qualitative visualizations were used to assess each stage.

\subsection{Quantitative Evaluation of Pix2Pix Preprocessing}

\begin{figure}[!h]    
    \centering
    \includegraphics[scale=0.2]{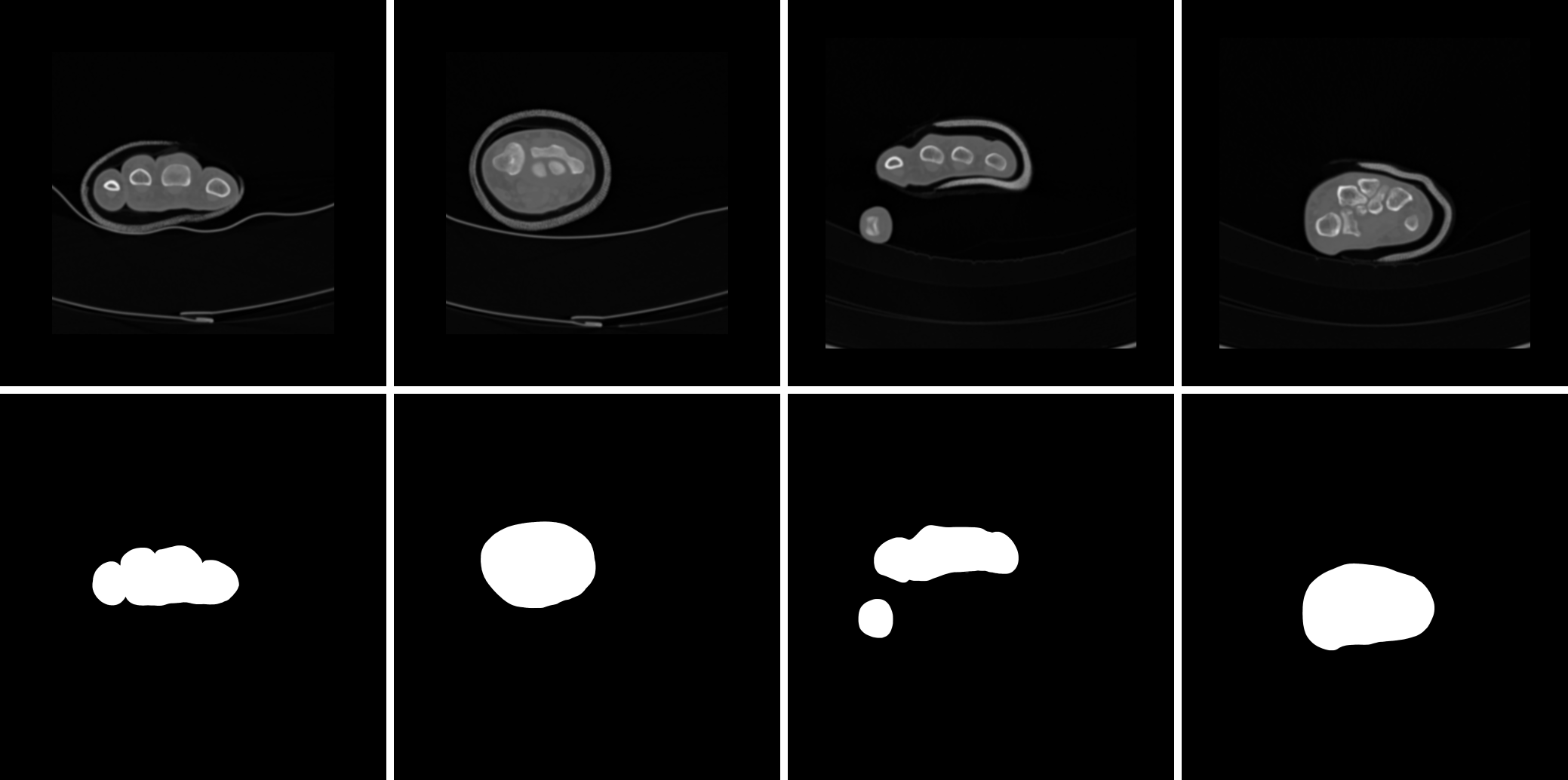}
    \caption{Example results of the Pix2Pix preprocessing stage. Top row: Original axial CT slices including background and plaster artifacts. Bottom row: Corresponding Pix2Pix outputs with artifacts removed, highlighting the preserved anatomical structures.}
    \label{fig1}
\end{figure}

Figure~\ref{fig1} illustrates representative results of the Pix2Pix preprocessing stage. The top row shows original axial CT slices containing anatomical structures alongside background and plaster artifacts. The bottom row shows the corresponding model outputs, demonstrating effective artifact removal while preserving anatomical integrity.

The Pix2Pix model achieved a mean Dice coefficient of 0.9856 and an IoU of 0.9720, indicating very high overlap between predicted and ground-truth masks. Precision and recall reached 0.9844 and 0.9872 respectively, demonstrating accurate identification of anatomical structures with a low false positive rate. Pixel-wise accuracy reached 0.9988. All reported values represent mean $\pm$ standard deviation computed across the held-out test set. Full quantitative results are summarized in Table~\ref{tab:pix2pix_metrics}.

\begin{table}[h]
\centering
\caption{Quantitative performance of the Pix2Pix preprocessing model on the held-out test dataset.}
\begin{tabular}{lc}
\hline
\textbf{Metric} & \textbf{Value (Mean $\pm$ SD)} \\
\hline
Dice Coefficient              & $0.9856 \pm 0.0124$ \\
Intersection over Union (IoU) & $0.9720 \pm 0.0236$ \\
Precision                     & $0.9844 \pm 0.0142$ \\
Recall                        & $0.9872 \pm 0.0201$ \\
Pixel-wise Accuracy           & $0.9988 \pm 0.0008$ \\
\hline
\end{tabular}
\label{tab:pix2pix_metrics}
\end{table}

\subsection{Bone and Skin Mesh Extraction in 3D Slicer}

The 3D Slicer-based workflow enabled extraction of anatomically coherent skin and bone mesh models from the Pix2Pix-cleaned CT volumes. The external soft-tissue envelope was recovered by boundary-connected air exclusion, while osseous structures were isolated by intensity thresholding and connected-component filtering. The mesh extraction workflow was applied to all 1,271 CT volumes. In the vast majority of cases, anatomically consistent meshes were generated without requiring manual correction, demonstrating the robustness of the Pix2Pix preprocessing stage in facilitating reliable downstream mesh generation. The resulting meshes preserved fine anatomical structures of the hand and distal forearm, providing reliable geometric inputs for skeletal fitting and statistical analysis.

\subsection{Skinning Quality Assessment}

\begin{figure}[!h]    
    \centering
    \includegraphics[scale=0.2]{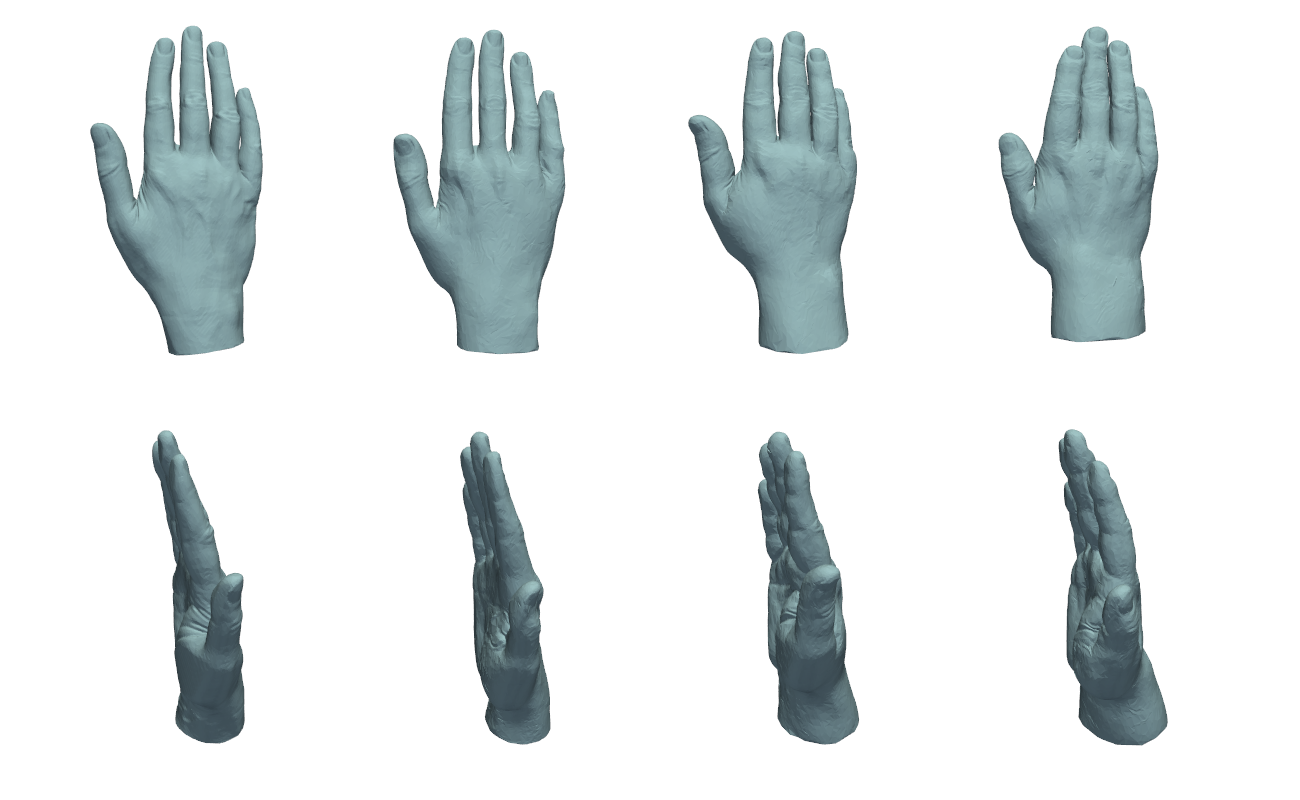}
    \caption{Visualization of pose-normalized hand models after implicit skinning across representative subjects. Top row: frontal views. Bottom row: corresponding lateral views. The consistent alignment across subjects demonstrates anatomically plausible pose normalization.}
    \label{fig2}
\end{figure}

Qualitative assessment of the implicit skinning stage was performed by visually inspecting the pose-normalized hand meshes across multiple subjects. As shown in Fig.~\ref{fig2}, the implicit skinning approach produced smooth anatomical deformations and maintained consistent joint geometry across subjects, particularly in highly articulated regions such as the wrist and finger bases. No visible candy-wrapper or collapsing artifacts — common failure modes of linear blend skinning — were observed in the normalized meshes, confirming the advantage of the implicit deformation formulation for anatomically complex geometries.

\subsection{Statistical Shape Analysis}

PCA-based statistical analysis was performed on 90 pose-normalized hand meshes. The first three principal components captured 82.6\% of the total anatomical variability: PC1 explained 44.0\%, PC2 explained 25.8\%, and PC3 explained 12.9\% of total variance. These components primarily correspond to variations in overall hand size (PC1), palm width and finger length proportions (PC2), and fine-scale shape asymmetries (PC3). The observed variability reflects true anatomical differences across subjects rather than residual pose inconsistencies, owing to the preceding GBCPD++-based non-rigid registration.

\begin{figure}[!h]    
    \centering
    \includegraphics[scale=0.15]{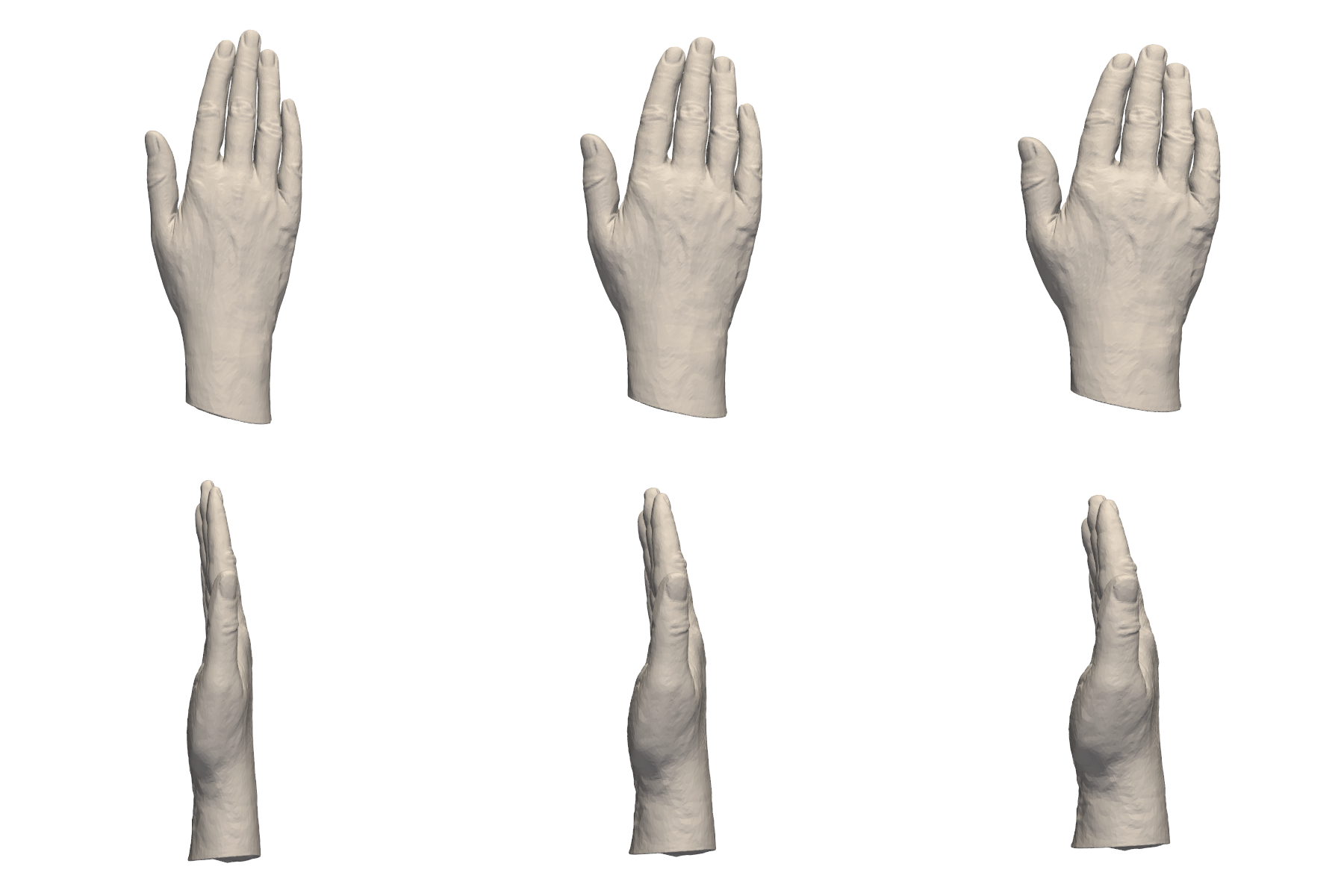}
    \caption{Representative hand shapes derived from the PCA-based statistical shape model. Top row: frontal views. Bottom row: lateral views. Left to right columns correspond to $-1$SD, mean, and $+1$SD along the first principal component, illustrating the primary mode of anatomical variation captured across the dataset.}
    \label{fig3}
\end{figure}

Figure~\ref{fig3} illustrates representative hand shapes generated by the statistical model, showing frontal and lateral views across the 
PCA-derived shape space.

Quantitative validation against the ANSUR~II dataset~\cite{gordon2015ansur} is presented in 
Table~\ref{tab:anthropometric_comparison}. The mean shape measurements derived from the proposed model show strong agreement with ANSUR~II normative values for hand length, palm length, hand circumference, and wrist circumference. A larger discrepancy was observed for hand breadth (78.7~mm vs.\ 85.0~mm), which is discussed in Section~\ref{sec:discussion}.

\begin{table}[h]
\centering
\caption{Comparison of hand anthropometric measurements between the proposed statistical model and the ANSUR~II dataset ($N = 6{,}068$; combined male and female). Proposed model range corresponds to the PCA-derived shape space ($\pm$1 SD along PC1).}
\label{tab:anthropometric_comparison}
\begin{tabular}{lccc}
\hline
\textbf{Measurement} & \textbf{Proposed} & \textbf{Proposed} 
& \textbf{ANSUR II} \\
 & \textbf{Mean (mm)} & \textbf{Range (mm)} 
 & \textbf{Mean $\pm$ SD (mm)} \\
\hline
Hand length         & 184.7 & [176.7,\ 190.6] & $189.3 \pm 11.5$ \\
Hand breadth        & 78.7  & [62.7,\ 94.5]   & $85.0  \pm 6.3$  \\
Hand circumference  & 209.4 & [168.7,\ 249.7] & $203.9 \pm 15.6$ \\
Palm length         & 115.0 & [111.0,\ 117.1] & $113.9 \pm 7.1$  \\
Wrist circumference & 170.3 & [133.4,\ 205.9] & $169.0 \pm 13.1$ \\
\hline
\end{tabular}
\end{table}

\section{Discussion}
\label{sec:discussion}

The experimental results demonstrate that the proposed AI-assisted pipeline provides a robust and scalable 
framework for reconstructing and analyzing hand anatomy from clinical CT data.

The Pix2Pix preprocessing stage achieved high segmentation accuracy (DSC = 0.9856, IoU = 0.9720), effectively removing plaster casts and background artifacts while preserving anatomical integrity. These results are comparable to state-of-the-art CT segmentation methods reported in the literature~\cite{wasserthal2022totalsegmentator, lindgren2021automated}, demonstrating that a cGAN-based approach is well-suited for artifact removal in clinical CT preprocessing.

The implicit skinning stage produced smooth, artifact-free pose normalization across subjects, consistently outperforming what would be expected from linear blend skinning in articulated regions such as the wrist and finger joints. This confirms the suitability of implicit deformation techniques for anatomically complex mesh alignment in medical imaging contexts.

The PCA-based statistical shape model demonstrated strong agreement with ANSUR~II normative measurements for four of the five evaluated dimensions. Hand length (184.7 vs.\ 189.3~mm), palm length (115.0 vs.\ 113.9~mm), hand circumference (209.4 vs.\ 203.9~mm), and wrist circumference (170.3 vs.\ 169.0~mm) showed differences within clinically acceptable ranges ($\leq$6~mm), supporting the anatomical validity of the reconstructed models. The observed discrepancy in hand breadth (78.7~mm vs.\ 85.0~mm) may be attributed to differences in measurement definitions between ANSUR~II and the landmark-based extraction used in this work, as well as demographic differences between the U.S. military population and the retrospective clinical cohort used in this study.

The dominance of the first principal component (44.0\%) is consistent with prior hand shape modeling studies, where overall hand size accounts for the largest source of inter-individual anatomical variation~\cite{handposture}. The cumulative variance of 82.6\% captured by the first three components demonstrates that the proposed pipeline produces a compact and interpretable statistical representation of hand morphology, comparable to state-of-the-art hand shape models reported in the literature~\cite{handposture, boss2022}.

The proposed framework has several potential clinical applications. Patient-specific statistical hand models can support the design of personalized orthotic and prosthetic devices by providing anatomically realistic reference geometries~\cite{orthoauto}. The generated models may also facilitate ergonomic product design through population-level morphological analysis, and may assist biomechanical simulation and surgical planning where accurate hand anatomy representation is required.

Several limitations should be acknowledged. First, statistical shape modeling was restricted to the 90 scans meeting strict anatomical quality criteria for surface correspondence; this subset size reflects the clinical reality of retrospective CT data rather than a methodological constraint. Second, skeletal landmark definition requires manual operator input, which may introduce inter-operator variability. Third, the current pipeline addresses static anatomical pose normalization and does not model dynamic joint motion. Fourth, the ANSUR~II reference population consists of U.S. military personnel, which may not fully represent the demographic characteristics of clinical patient populations.

Future work will focus on automated landmark detection to reduce operator dependency, improved vertex--bone weighting strategies for more accurate skinning, integration of dynamic motion models for functional biomechanical analysis, and extension of the pipeline to larger and more demographically diverse clinical datasets.

\section{Conclusion}

This work presented an AI-assisted pipeline for segmentation, pose normalization, and statistical modeling of hand anatomy from elbow-to-hand CT scans. The proposed framework integrates Pix2Pix-based artifact removal, 3D Slicer-based skin and bone mesh extraction, implicit skinning for anatomically consistent pose normalization, and PCA-based statistical shape modeling within a unified pipeline.

The Pix2Pix preprocessing stage achieved high segmentation accuracy on clinical CT data containing plaster casts and background artifacts. The implicit skinning approach produced smooth, artifact-free pose normalization across subjects. The resulting statistical shape model, capturing 82.6\% of anatomical variability within the first three principal components, showed strong agreement with ANSUR~II normative anthropometric measurements, supporting the anatomical validity of the reconstructed hand geometries and demonstrating that the pipeline can reliably capture population-level shape variability from retrospective clinical data.

The developed methodology provides a promising foundation for applications in personalized prosthetic and orthotic device design, ergonomic product development, and biomechanical simulation. Future research will focus on automated skeletal landmark detection, dynamic anatomical modeling, and validation on larger and more diverse clinical cohorts.

\end{document}